\documentclass{article}

    \PassOptionsToPackage{numbers, compress}{natbib}

\usepackage[final]{neurips_2024}

\usepackage[utf8]{inputenc} 
\usepackage[T1]{fontenc}    
\usepackage{hyperref}       
\usepackage{url}            
\usepackage{booktabs}       
\usepackage{amsfonts}       
\usepackage{nicefrac}       
\usepackage{microtype}      
\usepackage{xcolor}         
\usepackage{algorithm}
\usepackage{algpseudocode}%

\usepackage{amsmath}
\usepackage{amsfonts}
\usepackage{bm}

\def\eqref#1{equation~\ref{#1}}

\def\1{\bm{1}}

\def\vg{{\bm{g}}}

\def\vs{{\bm{s}}}

\def\vx{{\bm{x}}}

\DeclareMathAlphabet{\mathsfit}{\encodingdefault}{\sfdefault}{m}{sl}
\SetMathAlphabet{\mathsfit}{bold}{\encodingdefault}{\sfdefault}{bx}{n}

\def\gL{{\mathcal{L}}}

\def\gN{{\mathcal{N}}}

\DeclareMathOperator*{\argmax}{arg\,max}

\usepackage{graphicx} 
\usepackage{enumitem}  
\usepackage{xspace}  
\usepackage{amssymb}
\usepackage{pifont}

\definecolor{tabvline}{HTML}{a8a495}
\definecolor{prompt_blue}{HTML}{1f78b4}
\definecolor{prompt_red}{HTML}{d45c43}

\definecolor{green_im}{rgb}{0.0, 0.5, 0.0}

\title{DiP-GO: A \underline{Di}ffusion \underline{P}runer via Few-step \underline{G}radient \underline{O}ptimization}

\author{%
 Haowei Zhu\thanks{Work done during an internship at AMD.}$^{~~12}$, Dehua Tang$^{1}$, Ji Liu$^{1}$, Mingjie Lu$^{1}$, Jintu Zheng$^{1}$, Jinzhang Peng$^{1}$, Dong Li$^{1}$, \\ \textbf{Yu Wang$^{1}$, Fan Jiang$^{1}$, Lu Tian$^{1}$, Spandan Tiwari$^{1}$, Ashish Sirasao$^{1}$, Junhai Yong$^{2}$,} \\
  \textbf{Bin Wang\thanks{Corresponding author.}$^{~~2}$, Emad Barsoum\footnotemark[2]$^{~~1}$} 
  \\
  $^1$Advanced Micro Devices, Inc. ~~~~$^2$Tsinghua University
  \\
  \texttt{zhuhw23@mails.tsinghua.edu.cn;}  \\
  \texttt{\{dehua.tang, ji.liu,  jinz.peng, d.li, lu.tian, ebarsoum\}@amd.com;}\\
  \texttt{\{yongjh, wangbins\}@tsinghua.edu.cn} \\
}

\begin{document}

\maketitle
\begin{abstract}
Diffusion models have achieved remarkable progress in the field of image generation due to their outstanding capabilities. However, these models require substantial computing resources because of the multi-step denoising process during inference. While traditional pruning methods have been employed to optimize these models, the retraining process necessitates large-scale training datasets and extensive computational costs to maintain generalization ability, making it neither convenient nor efficient. Recent studies attempt to utilize the similarity of features across adjacent denoising stages to reduce computational costs through simple and static strategies. However, these strategies cannot fully harness the potential of the similar feature patterns across adjacent timesteps. In this work, we propose a novel pruning method that derives an efficient diffusion model via a more intelligent and differentiable pruner. At the core of our approach is casting the model pruning process into a SubNet search process. Specifically, we first introduce a SuperNet based on standard diffusion via adding some backup connections built upon the similar features. We then construct a plugin pruner network and design optimization losses to identify redundant computation. Finally, our method can identify an optimal SubNet through few-step gradient optimization and a simple post-processing procedure.
We conduct extensive experiments on various diffusion models including Stable Diffusion series and DiTs. Our DiP-GO approach achieves 4.4$\times$ speedup for SD-1.5 without any loss of accuracy, significantly outperforming the previous state-of-the-art methods.

\end{abstract}

\section{Introduction}
Diffusion models have undergone significant advancements over the past years due to the outstanding capabilities of diffusion probabilistic models (DPMs) \cite{ho2020denoising}. DPMs typically consist of two processes: the noise diffusion process and the reverse denoising process. Given their remarkable superiority in content generation, diffusion models have made significant progress in various fields of general image generation, including text-to-image generation \cite{rombach2022high, peebles2023scalable}, layout-to-image generation \cite{zhang2023adding, jia2024ssmg}, image editing \cite{chen2023anydoor, kawar2023imagic}, and image personalization \cite{ruiz2023dreambooth, kumari2023multi}. Furthermore, diffusion models have contributed to advancements in autonomous driving, ranging from driving dataset generation \cite{li2023drivingdiffusion, gao2023magicdrive, chen2023integrating} to perception model enhancement \cite{Chen2023PolyDiffuse, wang2023dire} through diffusion strategies.
However, DPMs often incur considerable computational overhead during both the training and inference phases. The high cost of inference, due to the multi-step denoising computation during the sampling process, can significantly impact their practical application. Many efforts \cite{song2020denoising, kim2023bksdm, Lee2023koala} have been made to improve the efficiency of diffusion models, which can be broadly divided into two types of optimization: inference sampling optimization and model structural optimization.

Sampling optimization methods reduce the number of sampling steps for generation without compromising image quality. For instance, DDIM \cite{song2020denoising} reduces these steps by exploring a non-Markovian process without requiring model retraining. LCM \cite{luo2023latent, luo2023lcmlora} enable image generation in fewer steps with retraining requirements. Structural optimization methods \cite{kim2023bksdm, Lee2023koala,fang2023structural,zhang2024laptopdiff} aim to reduce computational overhead through efficient model design and model pruning.
These methods require retraining the diffusion model, which entails significant computational overhead and large-scale datasets, making them neither convenient nor efficient.
DeepCache \cite{ma2023deepcache} proposes a novel training-free paradigm based on the U-Net architecture in diffusion models, caching and retrieving features across adjacent denoising stages to reduce redundant computation costs. However, DeepCache only reuses the output feature of a U-Net block in a denoising step via a simple and static strategy. We believe many intermediate features remain untapped, and the simple static strategy cannot fully exploit the potential of similar feature patterns across adjacent timesteps during inference, as observed in recent studies \cite{song2020denoising, luo2023latent, ma2023deepcache}.

To address these challenges, we introduce \textbf{Di}ffusion \textbf{P}runing via Few-step \textbf{G}radient \textbf{O}ptimization (DiP-GO), a method designed to achieve efficient model pruning with enhanced dynamism and intelligence. Our approach rethinks the diffusion model during inference by proposing a SuperNet based on standard diffusion via adding some backup connections built upon the similar
features, conceptualizing the inference process as a specific SubNet derived from our proposed SuperNet. We reformulate the diffusion model pruning into a SubNet search process.
By addressing the out-of-memory issue inherent in the backward process during expanded denoising timesteps using the gradient checkpoint~\cite{chen2016training} method, we introduce a plugin pruner that discovers an optimal SubNet surpassing existing methods through carefully designed optimization losses. Extensive experiments validate the effectiveness of our approach, demonstrating a 4.4$\times$ speedup on Stable Diffusion 1.5. Moreover, our method efficiently prunes the DiT model \cite{peebles2023scalable} without requiring retraining the diffusion model, achieving significant inference speedup.
Our contribution can be summarized as follows:
(1) We define a SuperNet based on standard diffusion and show how to obtain a SubNet. This transforms the diffusion optimization problem into an efficient SubNet search process without the need for retraining pretrained diffusion models.
(2) We design a plugin pruner tailored specifically for diffusion models. This pruner optimizes pruning constraints while maximizing synthesis capability. Additionally, we introduce a post-processing method for the pruner to ensure that the SubNet meets specified pruning requirements.
(3) We conduct extensive pruning experiments across various diffusion models, including Stable Diffusion 1.5, Stable Diffusion 2.1, Stable Diffusion XL, and DiT. Extensive experiments demonstrate the superiority of our method, achieving a notable 4.4 $\times$ speedup during inference on Stable Diffusion 1.5 without the need for retraining the diffusion model.


\section{Related Work}
\subsection{Efficient Diffusion Models}
The diffusion models, celebrated for their iterative denoising process during inference, play a pivotal role in content generation but are often hindered by time-consuming operations. To mitigate this challenge, extensive research has focused on accelerating diffusion models. Acceleration efforts typically approach the problem from two primary perspectives:

\textbf{Efficient Sampling Methods.}
Recent works focus on reducing the number of denoising steps required for content generation. DDIM \cite{song2020denoising} achieves this by exploring a non-Markovian process related to neural ODEs. Fast high-order solvers \cite{lu2022dpm, lu2023dpmsolver} for diffusion ordinary differential equations also enhance sampling speed. LCMs \cite{luo2023latent, luo2023lcmlora} treat the reverse diffusion process as an augmented probability flow ODE (PF-ODE) problem, inspired by Consistency Models (CMs) \cite{song2023consistency}, enabling generation in fewer steps. PNDM \cite{liu2022pseudo} emphasizes efficient sampling without retraining diffusion model. Additionally, ADD \cite{sauer2023adversarial} combines adversarial training and score distillation to transform pretrained diffusion models into high-fidelity image generators using only single sampling steps.

\textbf{Efficient Structural Methods.}
Other efforts concentrate on reducing the computational overhead associated with each denoising step. Previous methods \cite{kim2023bksdm, Lee2023koala, ma2023deepcache} have typically conducted extensive empirical studies to identify and remove non-critical layers from U-Net architectures to achieve faster networks.
BK-SDM \cite{kim2023bksdm} customizes three efficient U-Nets by strategically removing residual and attention blocks. Derived from BK-SDM, KOALA \cite{Lee2023koala} develops two efficient U-Nets of varying sizes tailored for SD-XL applications.
Diff-pruning \cite{fang2023structural} employs Taylor expansion over pruned timesteps to pinpoint essential layer weights, optimizing model efficiency without sacrificing performance.
DeepCache \cite{ma2023deepcache} enhances inference efficiency by reusing predictions from blocks in previous timesteps within the U-Net architecture.
LAPTOP-Diff \cite{zhang2024laptopdiff} tackles optimization problems with a one-shot pruning approach, incorporating normalized feature distillation to streamline retraining processes.
T-GATE \cite{tgate} not only reduces computation overhead but also marginally lowers FID scores by omitting text conditions during fidelity-improvement stages.

In addition to the two primary acceleration methods, other strategies such as distillation \cite{sauer2023adversarial, salimans2022progressive, berthelot2023tract}, early stopping \cite{lyu2022accelerating}, and quantization \cite{he2023ptqd} are commonly employed to enhance performance and efficiency. However, most of these strategies necessitate retraining pretrained models.
Our method falls under the category of efficient structural methods by focusing on reducing inference time at each timestep. Importantly, these efficiency gains are achieved without retraining the diffusion model.

\subsection{Model Optimization}
\textbf{Network Pruning.}
The taxonomy of pruning methodologies typically divides into two main categories: unstructured pruning methods \cite{park2020lookahead, dong2017learning, sanh2020movement} and structural pruning methods \cite{li2017pruning, elkerdawy2020filter, ding2019centripetal, liu2021group}. Unstructured pruning methods involve masking parameters without structural constraints by zeroing them out, often requiring specialized software or hardware accelerators. In contrast, structured pruning methods generally remove regular parameters or substructures from networks.
Recent works have been interested in accelerating transformers. Dynamic skipping blocks, which involve selectively removing layers while maintaining the overall structure, have emerged as a paradigm for transformer compression \cite{liu2024updp, zhang2020accelerating, dong2021attention, hou2020dynabert}. However, applying structural pruning techniques to diffusion modeling poses unique challenges that necessitate reevaluating conventional pruning methods.

\section{Methodology}

In this study, we introduce the Diffusion Pruner via Few-step Gradient Optimization (DiP-GO), which utilizes a neural network to predict whether to skip or keep each computational block during inference. Our primary objective is to identify the optimal subset of computational blocks that facilitate denoising with minimal computational overhead. As illustrated in Figure~\ref{fig:overview}, our method comprises three main components: a neural network pruner, optimization losses, and a post-process algorithm to derive the pruned model based on the predictions of pruner.
The neural network pruner is designed with learnable queries inspired by DETR \cite{carion2020end} to predict the state of each block. Our proposed optimization losses include sparsity and consistency constraints for generation quality, guiding the pruner to accurately assess the importance of each block.
In this Section, we first revisit the framework of diffusion models in Section \ref{sec:Preliminary}, emphasizing their potential for exploring pruned networks. In Section \ref{sec:supernet}, we introduce a SuperNet based on diffusion models and demonstrate how to derive a SubNet or pruned network from it for inference acceleration, highlighting the challenges in achieving an optimal SubNet.
Section \ref{sec:dip-go} details our method, including the neural network pruner, optimization losses, and post-process algorithm for obtaining a SubNet that meets pruning requirements. Finally, we provide insights into the training and inference processes of our method.

\subsection{Preliminary}
\label{sec:Preliminary}

We begin with a brief introduction to diffusion models. Diffusion models are structured to learn a series of sequential state transitions with the goal of iteratively refining random noise sampled from a known prior distribution towards a target distribution $x_0$ that matches the data distribution. During the forward diffusion process, the transition from $x_{t-1}$ to $x_t$ is initially determined by a forward transition function, which can be described as follows:

\begin{equation}\label{Eq:forward}
    q(x_t|x_{t-1}) = \mathcal{N}(x_t;\sqrt{1-\beta_t}x_{t-1}, \beta_t\mathbf{I})
\end{equation}
where the hyperparameter $\{\beta_t \in (0,1)\}_{t=1}^T$ increases with each successive time step $t$.

To generate samples from a learned diffusion model, it involves a series of reverse state transitions from $\vx_T \rightarrow \dots \rightarrow \vx_0$ to denoise random noise $\vx_T\sim\gN(\mathbf{0}, \mathbf{I})$ into the clean data point $\vx_0$. At each timestep, the denoised output $\vx_{t-1}$ is predicted by approximating the noise prediction network, which is conditioned on the time embedding $t$ and the previous data point $\vx_t$:

\begin{equation}\label{Eq:reverse}
    p_\theta(x_{t-1}|x_t) = \mathcal{N}(x_{t-1};\frac{1}{\sqrt{a_t}}(x_t-\frac{\beta_t}{\sqrt{1-\overline{a}_t}}z_\theta(x_t,t)),\beta\mathbf{I})
\end{equation}

where the covariance constant $\beta_t  = 1 - \alpha_t$, $\overline{a}_t = \prod_{i=1}^T \alpha_i$, and $z_\theta(x_t,t)$ are the parameterized deep neural networks. With the reverse Markov chain, we can iteratively sample from the learnable transition kernel $x_{t-1} \sim p_\theta(x_{t-1}|x_t)$ until $t=1$.

Diffusion modes typically require multi-step conditional sampling to gradually obtain the target sample point $x_0$. However, recent studies \cite{ song2020denoising, luo2023latent, ma2023deepcache} have highlighted that multi-step inference processes involve substantial redundant feature computations, particularly in noise prediction networks like UNet and Transformer. For example, in Stable Diffusion 1.4 models with 25 steps, Multiply-Accumulate Operations (MACs) of UNet can comprise up to 87.2\% of the total computational load \cite{kim2023bksdm}. This underscores significant potential for accelerating inference by effectively eliminating these redundancies. In this work, we propose accelerating the diffusion model by integrating a differentiable pruning network designed to identify and remove these redundant computations.

\subsection{SuperNet and SubNet of Diffusion Model}
\label{sec:supernet}

Our goal is to identify and remove unimportant blocks during inference to accelerate the process. To achieve this, we introduce a SuperNet based on the diffusion model. This SuperNet is designed to facilitate block removal while ensuring the pruned model maintains inference capability through additional connections. Our approach effectively eliminates unimportant blocks during inference, essentially deriving a SubNet from the SuperNet by skipping these unnecessary components. Thus, the pruning process can be conceptualized as a SubNet search within the SuperNet framework.

\begin{figure*}[ht]
\begin{center}
\centerline{\includegraphics[width=1.0\linewidth]{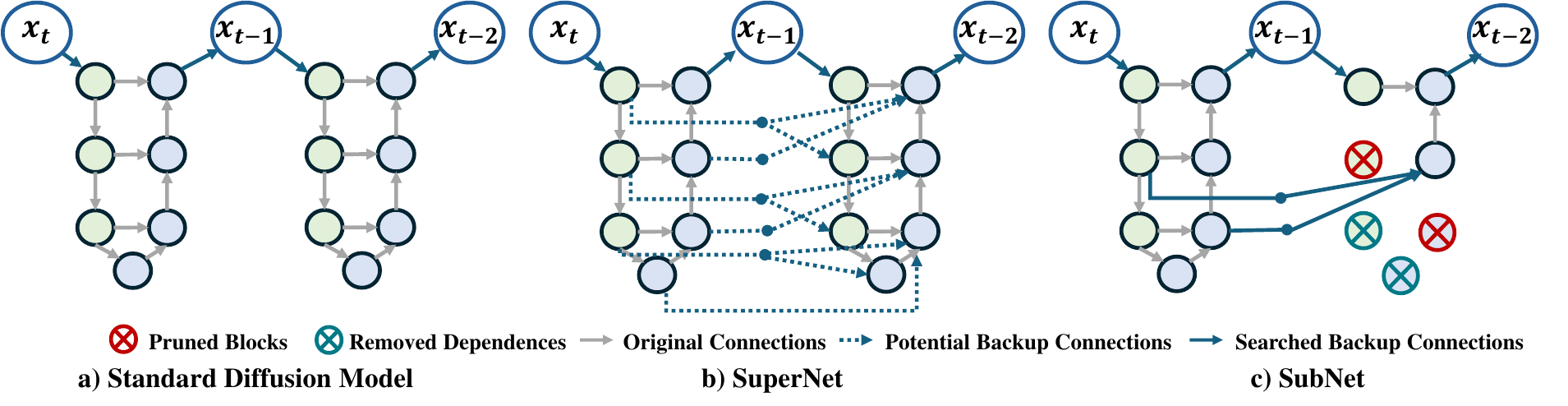}}
\caption{Overview of the SuperNet and SubNet. Standard diffusion models execute the full inference path step by step. In our framework, we propose a SuperNet based on the original flow and integrate backup connections to facilitate block removal. This allows the partial inference SubNet to efficiently eliminate redundant computational costs.}
\label{fig:supernet}
\end{center}
\vspace{-8mm}
\end{figure*}

\textbf{How to Construct a SuperNet.} 
Recent studies \cite{song2020denoising, luo2023latent,ma2023deepcache} have observed that diffusion models often exhibit similar feature patterns across adjacent timesteps during inference. Building on this insight, we enhance the standard diffusion model's inference phase by introducing additional connections from the current timestep to the previous one. These connections serve as backups for blocks that may be removed, ensuring each block retains valid inputs even if its dependent blocks are eliminated for acceleration. Specifically, for all inputs of each block across all timesteps except the inital step during inference, we establish a backup input connection to the corresponding block in the previous timestep, as illustrated in Figure~\ref{fig:supernet}.

\textbf{How to Obtain a SubNet.} 
To construct the SuperNet for the standard diffusion model, we introduce additional connections that ensure a valid SubNet selects either the original input connection or the backup input connection, but not both simultaneously. This design principle mandates that if a dependent block is pruned, its original input connection is also eliminated to reflect the block's removal. Conversely, if the dependent block is retained, the backup input connection is removed to maintain efficient inference, as depicted in Figure~\ref{fig:supernet}.

We draw inspiration from the Lottery Ticket Hypothesis (LTH) \cite{frankle2018lottery}, which posits the existence of a sub-network capable of achieving comparable performance to the original over-parameterized network for a given task, but with fewer unnecessary weights. Moreover, prior work \cite{ma2023deepcache} has explored manually removing redundant computations by caching features across adjacent steps. Thus, our approach seeks to identify an optimal SubNet from the SuperNet, maximizing diffusion model acceleration while minimizing any loss in generation quality.

\begin{figure}[t]
\begin{center}
\centerline{\includegraphics[width=1.0\linewidth]{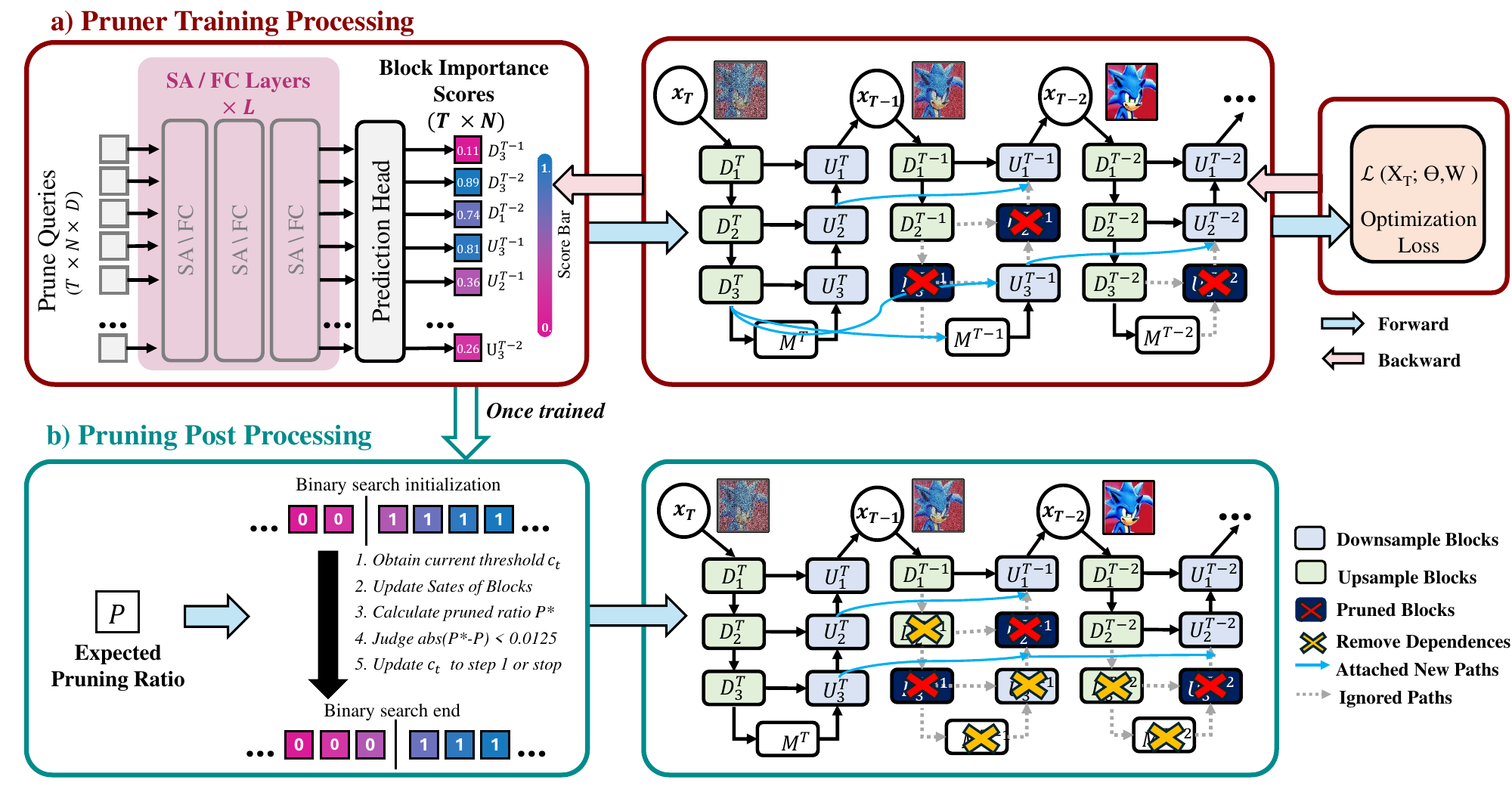}}
\caption{Overview of our diffusion pruner. a) DiP-GO employs a pruner network to learn the importance scores of blocks in the diffusion sampling process. It takes $N \times T$ queries as input and passes them through stacked self-attention (SA) and fully connected (FC) layers to capture the structural information in existing diffusion models. The network predicts the partial inference paths based on the $N \times T$ importance scores and is optimized by consistent and sparse loss. b) Once trained, the pruner network is discarded. We can infer the optimal partial inference path with expected computational costs via post-processing based on the predicted importance scores.}
\label{fig:overview}
\end{center}
\vspace{-8mm}
\end{figure}

\textbf{Hard to Obtain an Optimal SubNet.}
The challenge of obtaining an optimal SubNet is compounded by the large number of blocks expanded during inference. In a diffusion pipeline with $N \times T$ blocks (where $N$ is the number of blocks per timestep and $T$ is the number of timesteps), each block's decision to be kept or removed results in $2^{N \times T}$ possible configurations. For instance, a 50-step PLMS setup \cite{liu2022pseudo}, considering 9 blocks in the U-Net, yields $2^{450}$ choices (> $10^{135}$). Traditional search methods like random search and genetic algorithms \cite{holland1992genetic} often struggle in such vast search spaces. Gradient-based optimization offers a promising approach to tackle this challenge. However, there are significant hurdles to overcome. First, effectively modeling discrete block states (kept or removed) with parametric methods poses difficulties. Second, training the entire model, comprising both the parametric model and the expanded diffusion model with denoising timesteps, risks encountering out-of-memory (OOM) issues.

\vspace{-10pt}
\subsection{Our DiP-GO Approach}
\label{sec:dip-go}

In this study, we introduce a diffusion pruner network designed to predict importance scores for all blocks during reverse sampling as depicted in Figure \ref{fig:overview}. To optimize the pruner network effectively, we employ two key optimization losses: consistency and sparsity losses, leveraging few-step gradient optimization. Addressing the OOM issue inherent in such computations, we implement gradient checkpointing and half-precision floating-point representation techniques, enabling efficient search processes on a single GPU.
Once the pruner network trained, we extract predicted importance scores for all blocks. Subsequently, we devise a post-processing algorithm to utilize these scores, generating pruned SubNets of diffusion models that satisfy specific pruning criteria.

\textbf{Pruner Network.}
Our pruner network comprises three main components:  $N \times T$ learnable queries, a query encoder, and a prediction head. We design the learnable queries to match the number of all blocks during inference. These queries are optimized with sparsity and consistency loss constraints to learn the contextual information necessary for predicting the importance score of each block.
For the query encoder, we provide two options: a simple version with several stacked linear layers, and a more complex version with several stacked self-attention layers to facilitate interaction among the learnable queries. Our experiments demonstrate that both versions can effectively obtain optimal SubNets in various diffusion models under different pruning requirements.
The prediction head consists of $N \times T$ simple branches, each containing two stacked linear layers followed by a softmax operation. The final linear layer has a dimension of 2, and the softmax output represents the importance scores of a block. During training or inference, the query embeddings are transformed into output embeddings via the query encoder. These embeddings are then independently decoded into binary vectors by the multi-layer prediction head, resulting in $N \times T$ importance scores for all blocks.

\textbf{Optimization Losses.}
The $k$-th predicted binary vector of importance score, denoted as $\vs^k$, represents the likelihood of its corresponding block being removed or kept in the denoising process. A gate 
$\vg \in \{0, 1\}^{TN}$ is derived based on $\vs$, where $\vg^k=0$ or $\vg^k=1$ indicate removing or keeping the $k$-th computation block, respectively. Only the blocks that are kept according to $\vg$ will be calculated in the denoising process. 
However, directly converting predicted probabilities $\vs$ into discrete gates $\vg$ with $\argmax$ is non-differentiable. To address this issue, we utilize the Straight-Through (ST) Estimator \cite{jang2016categorical} to approximate the real gradient $\nabla_\theta \vg$ with the gradient of the soft prediction $\nabla_\theta \vs $.
To encourage both high-fidelity predictions and minimal computation block usage, we design our training objective function as a combination of consistent loss $\gL_c$ and sparse loss $\gL_s$, formulated as follows:
\begin{equation}
    \gL(\vx_T;\theta,W) =\gL_c+ \alpha_s \gL_s \\
= \begin{cases}
    f(\vx_0^p, \vx_0^{gt} ) +  \frac{\alpha_s}{NT}\sum_k^{NT}\gamma^k \vg^k & \text{if } sparsity < \tau \\
    f(\vx_0^p, \vx_0^{gt} ) & \text{if } sparsity \geq \tau
\end{cases}
\label{Eq:loss}
\end{equation}

Here, $\alpha_s$ represents a hyperparameter used to balance the consistent and sparse losses. $\theta$ and $W$ denote the pruner network and pretrained diffusion model, respectively. $f(\cdot)$ denotes a distance function that evaluates the consistency between the generated clean data point $\mathbf{x}_0^p$ from partial inference of the pruned SubNet and the $\mathbf{x}_0^{gt}$ from full inference. This function can be any distance measure, and in this work, we utilize a negative SSIM loss \cite{wang2004image}. The sparse loss encourages minimal computational usage and is weighted by the computational flops proportion $\gamma^k$ of the $k$-th block, thereby imposing a greater penalty on heavier blocks. The calculation of $\gamma$ takes into account the cascading relationships between blocks. Specifically, when a block is pruned, the associated dependent blocks will also pruned. Therefore, the flops reduction from pruning a block includes the block itself and its dependent blocks. We denote the flops reduction ratio after pruning the $k$-th block as $\gamma^k$. The flops ratio $\gamma$ is in the range $[0, 1]$.
The sparse loss is only introduced when the sparsity (pruning ratio) is below a certain threshold $\tau$. This compound loss controls the trade-off between efficiency (block usage) and accuracy (generation quality). 

\textbf{Post-Processing Algorithm.} After training the pruner network, our diffusion pruner is able to predict which computation blocks during inference contribute less to generation quality based on the importance scores for all the blocks. As the importance scores are continuous values in inference phase, they can not be utilized directly to identify which blocks should be removed to meet given pruning requirements. Therefore, we present a post-process algorithm to obtain an appropriate threshold for these importance score to meet the pruning requirements as shown in  Algorithm \ref{alg:post-p} in Appendix~\ref{Append_B}.
Considering the required pruning sparsity, we use bisection lookup to select the appropriate threshold value to identify which blocks should be removed to meet the pruning ratio. Specifically, the blocks whose important scores below the threshold should be removed and the kept blocks should update their input connections as mentioned in Section \ref{sec:supernet} to maintain the pruned model inference. Thus a pruned model met the pruning ratio has been obtained.

\textbf{Training and Inference Details.}
In the training phase, the prompt inputs are fed into the diffusion model to obtain two kinds of outputs, one is generated by the baseline diffusion model and the other is generated by the pruned model obtained via the current predictions of the pruner network. Then our proposed losses are utilized to optimize the pruner network to enable distinguishing the less important blocks.
In the pruner's network, we initialize the weight of the last linear layer's output channel to 0 and its bias to 1. This setup ensures that at the beginning of training, the consistency loss is 0 and the sparsity loss is 1, facilitating smooth training. As training progresses, the sparsity loss gradually decreases while the number of pruned blocks increases, causing the consistency loss to rise. To maintain network fidelity after pruning, we switch to training only with the consistency loss once the sparsity loss reaches 0.2, continuing until training is complete.
Once the pruner is well trained, we can obtain pruned models to meet the pruning requirements via our post-process algorithm.

\section{Experiments}

\vspace{-5pt}
\subsection{Experimental Setup}
\label{sec: Experimental setup}
\vspace{-5pt}
\textbf{Pre-trained Model and Datasets.}
We select four official pretrained Diffusion Models (i.e., SD-1.5 \cite{rombach2022high}, SD-2.1 \cite{rombach2022high}, SD-XL \cite{podell2023sdxl} and DiT \cite{peebles2023scalable}) to evaluate our approach. 
The SD series models are constructed on the U-Net \cite{ronneberger2015u} and the DiT is constructed on the transformer \cite{vaswani2017attention}.
We utilize a subset of the DiffusionDB \cite{wangDiffusionDBLargescalePrompt2022} dataset comprising 1000 samples to train our pruner network, utilizing only textual prompts. 
Following previous works \cite{tgate, ma2023deepcache}, we evaluate the DiP-GO on three public datasets, i.e., PartiPrompts \cite{yu2022scaling}, MS-COCO 2017 \cite{lin2014microsoft} and ImageNet \cite{deng2009imagenet}.

\textbf{Evaluation Metrics.}
We employ the Fréchet Inception Distance (FID) \cite{heusel2017gans} metrics to assess the quality of images created by the generative models. FID quantifies the dissimilarity between the Gaussian distributions of synthetic and real images. A lower FID score indicates a closer resemblance to real images in the generative model. 
Additionally, we utilize the CLIP Score \cite{hessel2021clipscore} (ViT-g/14) to evaluate the relational compatibility between images and text.

\textbf{Implementation Details.}
For Stable Diffusion models, we utilize the SGD optimizer with a cosine learning schedule for 1000 steps of training. The batch size, learning rate, and weight decay are set to 1, 0.1, and $1\times10^{-4}$, respectively. The hyperparameters $\alpha_s$, $\tau$, and the query embedding dimension $D$, along with the encoder layer number $L$, are set to 1, 0.2, 512, and 1, respectively. For the Diffusion Transformer model, we use the same experimental configuration as for the stable diffusion model, except that the learning rate set to 10.
To evaluate the inference efficiency, we evaluate the Multiply Accumulate Calculation (MACs), Parameters (Params), and Speedup for all models with batch size of 1 in the PyTorch 2.1 environment on the AMD MI250 platform.
Besides, we report MACs in those tables, which refer to the totals MACs for all steps.

\begin{table}[h]
\centering
\caption{Comparison with PLMS, BK-SDM and DeepCache on SD-1.5. We utilize prompts in PartiPrompt and COCO2017 validation set.}
\label{tbl:main_stable_diffusion}
\small
\resizebox{\linewidth}{!}{
\begin{tabular}{l | c | c c c | c c c}
  \toprule
  & & \multicolumn{3}{c|}{\bf PartiPrompts} & \multicolumn{3}{c}{\bf COCO2017}\\
  \bf Method & \bf Pruning Type &  \bf MACs $\downarrow$ & \bf Speedup $\uparrow$ & \bf CLIP Score $\uparrow$ & \bf MACs $\downarrow$ & \bf Speedup $\uparrow$ & \bf CLIP Score $\uparrow$ \\
  \midrule
  
  PLMS - 50 steps & Baseline & 16.94T &  1.00$\times$ & 29.51 & 16.94T & 1.00$\times$ & 30.30\\
  \midrule
  BK-SDM - Base & Structured   & 11.19T & 1.49$\times$ & 28.88 & 11.19T & 1.45$\times$ & 29.47 \\
  
  PLMS - 25 steps & Fast Sampler & 8.47T & 2.04$\times$ & 29.33 & 8.47T & 1.91$\times$ & 29.99\\
  PLMS - Skip - Interval=2  & Structured & 8.47T & 2.04$\times$ & 19.74 & 8.47T & 1.91$\times$ & 16.78\\
  DeepCache & Structured & 6.52T & 2.15$\times$ & 29.46 & 6.52T & 2.11$\times$ & 30.23 \\

  \bf Ours (w/ Pruned-0.80) & Structured & \bf3.38T & \bf4.43$\times$ & \bf29.51 & \bf3.38T & \bf4.40$\times$ & \bf30.29 \\ 

  \midrule
  BK-SDM - Small & Structured   & 10.88T & 1.75$\times$ & 27.94 & 10.88T & 1.68$\times$ & 27.96 \\

  PLMS - 15 steps & Fast Sampler & 5.08T & 2.89$\times$ & 28.58 & 5.08T  & 2.59$\times$ & 29.39\\
  \bf Ours (w/ Pruned-0.85) & Structured & \bf 2.54T & \bf5.52$\times$ & \bf29.07 & \bf2.54T & \bf5.46$\times$ & \bf29.84 \\
  \bottomrule
\end{tabular}
}
\vspace{-2mm}
\end{table}

\begin{table*}[h]
\centering
\vspace{-4mm}
\caption{Comparison of computational complexity,  inference speed, CLIP Score and FID on the MS-COCO 2017 validation set on SD-2.1.}
\vspace{2mm}
\resizebox{0.7\linewidth}{!}{
\begin{tabular}{lcccc}
\toprule
\textbf{Inference Method} & \textbf{MACs$\downarrow$} & \textbf{Speedup}$\uparrow$ & \textbf{CLIP Score $\uparrow$} & \textbf{FID-5K $\downarrow$} \\ 
\midrule
SD-2.1-50 steps \cite{rombach2022high}         & 38.04T       &  $1.00\times$         & 31.55  & 27.29                     \\ 
\midrule 

SD-2.1-20 steps \cite{rombach2022high}         & 15.21T       &  $2.49\times$        & \bf31.53  & 27.83                     \\ 
\bf Ours (w/ Pruned-0.7)    & 11.42T    & $3.02\times$      & 31.50  & \bf25.98                     \\ 
\bf Ours (w/ Pruned-0.8)    & 7.61T       & $3.81\times$       & 30.92    & 27.69                \\ 
\bottomrule
\end{tabular}
}
\vspace{-4mm}
\label{tab:sd2.1}
\end{table*}

\subsection{Comparison with State-of-the-Art Methods on Different Base Models}

\textbf{Stable Diffusion on PartiPrompt and COCO2017.}
We compare our method with the state-of-the-art (SOTA) compression methods on Stable Diffusion 1.5 (SD-1.5), and the results are summarized in Table~\ref{tbl:main_stable_diffusion}. Compared to the SOTA DeepCache~\cite{ma2023deepcache}, our approach demonstrates significant performance improvements, achieving nearly $2\times$ fewer MACs while maintaining better CLIP Scores. Our method can achieve $4.4\times$ speedup compared to the baseline model. Furthermore, our method does not require training the diffusion model, which preserves the pre-trained knowledge of the diffusion model. Also, we apply our method on the SD-2.1 model to verify the effectiveness, as shown in Table~\ref{tab:sd2.1} , our method achieves significant acceleration while maintaining generation quality, demonstrating its superiority.

\textbf{Diffusion Transformers on ImageNet.}
To the best of our knowledge, we are the first to apply pruning to DiT \cite{peebles2023scalable} model. Therefore, we have replicated a training-free acceleration method, DeepCache with intervals = 2 and 5, on DiT for comparison. The results in Table~\ref{tab:dit} show that our method can speed up the original DiT model by a factor of 2.4 with minimal performance loss, while DeepCache has a lower speedup ratio when applied to the DiT model. This can be attributed to DeepCache's overreliance on pre-defined structures, whereas our method can automatically learn the optimal pruning strategy for the given model, thereby achieving superior performance.

\begin{table*}[t]
\centering
\caption{Comparison of pruning type, computational complexity, FID and inference speed on the ImageNet validation datasets on DiT. * denotes the results reproduced with diffusers \cite{diffusers}.}
\vspace{10pt}
\label{tab:dit}
    \small
    \resizebox{0.8\linewidth}{!}{
        \begin{tabular}{c|c|c|c|c}
        \toprule  
       \bf Method & \bf Pruning Type & \bf MACs $\downarrow$ & \bf FID-50K $\downarrow$ & \bf Speedup $\uparrow$ \\
        \midrule  
        DiT-XL/2-250 steps & - & 29.66T & 2.27 &  $1.00 \times$\\
        \midrule  
        DiT-XL/2*-250 steps & Baseline & 29.66T & 2.97 &  $1.00 \times$\\
        \midrule  
        DiT-XL/2*-110 steps & Fast Sampler & 13.05T & 3.06 & $2.13 \times$\\
        DiT-XL/2*-100 steps & Fast Sampler & 11.86T & 3.17 & $2.46 \times$\\
        DeepCache(DiT-XL/2*)-N=2 & Structured Pruning & 15.88T & 3.07 &  $1.76 \times$ \\
        \textbf{Ours (DiT-XL/2* w/ Pruned-0.6)} & Structured Pruning & 11.86T & \bf3.01 &  $ \bf2.43 \times$ \\
        \midrule  
        DiT-XL/2*-70 steps & Fast Sampler & 8.30T & 3.35 & $3.49 \times$\\
        DeepCache(DiT-XL/2*)-N=5 & Structured Pruning & 6.77T & 3.20 &  $3.44 \times$ \\
        \textbf{Ours (DiT-XL/2* w/ Pruned-0.75)} & Structured Pruning& 7.40T & \bf3.14 &  $\bf3.60 \times$ \\
        \bottomrule 
        \end{tabular}
     }
     \vspace{-4mm}
\end{table*}

\subsection{Compatibility with Fast Sampler}
We investigate the compatibility of DiP-GO with methods that prioritize reducing sampling steps using faster samplers: DDIM \cite{song2020denoising}, DPM-Solver \cite{lu2023dpmsolver}, and LCM \cite{luo2023latent}. As shown in Table \ref{tab:sampler}, it indicates that our method further improves computational efficiency on existing fast samplers. Specifically, we reduce MACs by a factor of 5 on the SD-1.5 with DDIM sampler and by $3.36 \times$ on the SD-2.1 with DPM-Solver. Our method achieves nearly unchanged performance with significant acceleration. 
Additionally, our method benefits from information redundancy in multi-step optimization processes, showing relatively limited acceleration performance on fewer-step LCM due to its low redundancy in features across adjacent timesteps.

\begin{table}[h]
\centering
\caption{Comparison with PLMS, SSIM, and LCM samplers. We evaluate the effectiveness of our methods on COCO2017 validation set.}
\resizebox{0.7\linewidth}{!}{
\begin{tabular}{@{}lcccc@{}}
\toprule
\textbf{Sampler} & \multicolumn{2}{c}{\textbf{Base Model}} & \multicolumn{2}{c}{\textbf{Ours}} \\
\cmidrule(lr){2-3} \cmidrule(lr){4-5}
                  & \textbf{MACs $\downarrow$} & \textbf{CLIP Score $\uparrow$} & \textbf{MACs $\downarrow$} & \textbf{CLIP Score $\uparrow$} \\ \midrule
DDIM (SD-1.5 w/ 50 steps)  & 16.94T & 30.30 & 3.38T & 30.29 \\
DPM (SD-2.1 w/ 50 steps)  & 38.04T & 31.55 & 11.42T & 31.50 \\
DPM (SD-2.1 w/ 25 steps)  & 19.02 T & 31.59 & 9.51T & 31.52 \\
LCM (SD-XL w/ 4 steps)       & 11.95T & 31.92 & 11.58T & 31.30 \\ \bottomrule
\end{tabular}
}
\vspace{-4mm}
\label{tab:sampler}
\end{table}

\subsection{Ablation Study}
\label{sec: ablation}

\textbf{Compared with Different Consistent Constraints.} 
We further compare other alternatives explored for our consistent loss designs, we further scrutinize additional options, including L1, L2, SSIM, and L1+SSIM losses, as depicted in Table \ref{tab:loss}. The results demonstrate that SSIM emerges as the most effective choice, boasting the highest CLIP-Score. In contrast, the L1 loss function often results in image blurring or distortion due to its sensitivity to pixel-level differences, the L2 loss may yield overly smoothed images by penalizing squared differences between pixels. Conversely, the combination of L1+SSIM loss attempts to address these limitations but can complicate the training process and suffer from trade-offs. Therefore, SSIM emerges as the preferred choice in our consistent loss designs, offering superior accuracy and stability while preserving image quality.

\begin{table*}[ht]
\centering
\vspace{-4mm}
\caption{Comparison with different consistent loss types. Here we conduct pruning experiments with 80\% sparsity on COCO2017 validation using SD-1.5.}
\vspace{10pt}
\resizebox{0.5\linewidth}{!}{
\begin{tabular}{c|cccc}
\toprule
\bf Loss Type   & \bf L1 & \bf L2 & \bf SSIM & \bf L1 + SSIM \\\midrule
CLIP Score$\uparrow$  & 29.94  & 29.71  & \bf 30.29    & 29.77          \\ \bottomrule
\end{tabular}
}
\vspace{-2mm}
\label{tab:loss}
\end{table*}

\textbf{Effect of Gradient Optimization.}
As the traditional search algorithm can also obtain SubNets from our proposed SuperNet.
It is crucial to validate whether traditional search-based algorithms yield positive effectiveness. We assess two search algorithms: random search and genetic algorithm-based search \cite{holland1992genetic} in Table \ref{tab:search}. We have iterated the search 1000 times using the first 500 images of the test set as a calibration dataset. Remarkably, we observe that the search time of traditional search algorithms is significantly longer than the training time of our method due to a large number of evaluations. Moreover, due to the vast search space, traditional search algorithms struggle to achieve satisfactory results. Additionally, traditional search algorithms lack the “once-for-all” characteristic, requiring re-execution when faced with deployment scenarios demanding different computational resources. In contrast, leveraging the parametric pruner network, our method achieves superior performance with reduced running time and is more adaptable to diverse development scenarios.

\begin{table*}[ht]
\centering
\vspace{-4mm}
\caption{Comparison of cost time, computational complexity and CLIP-Score between Random Search and GA search strategies on Stable Diffusion 1.5.}
\vspace{2mm}
\resizebox{0.7\linewidth}{!}{
\begin{tabular}{c|c|c|c|c}
\toprule
\textbf{Method} & \textbf{Cost GPU Hours $\downarrow$} & \textbf{Pruning Ratio} & \textbf{MACs $\downarrow$} & \textbf{CLIP Score $\uparrow$} \\ 
\midrule
PLMS-50 steps & - & - & 16.94T & 30.30 \\ 
\midrule
Random Search & 25 & 0.80 & 2.96T & 28.73 \\ 
GA Search & 25 & 0.80 & 3.34T & 29.37 \\ 
\bf Ours & 2.3 & 0.80 & 3.38T & \bf 30.29 \\ 
\midrule
Random Search & 24  & 0.85 & 2.90T & 27.22 \\ 
GA Search & 24  & 0.85 & 2.73T & 28.61 \\ 
\bf Ours & 2.2 & 0.85 & 2.54T & \bf 29.84 \\ 
\midrule
Random Search & 23 & 0.90 & 1.94T & 24.07 \\ 
GA Search & 23 & 0.90 & 2.04T & 25.14 \\ 
\bf Ours & 2.2 & 0.90 & 1.69T & \bf 28.72 \\ 
\bottomrule
\end{tabular}
}
\vspace{-4mm}
\label{tab:search}
\end{table*}

\begin{figure*}[ht]
\begin{center}
\centerline{\includegraphics[width=1.0\linewidth]{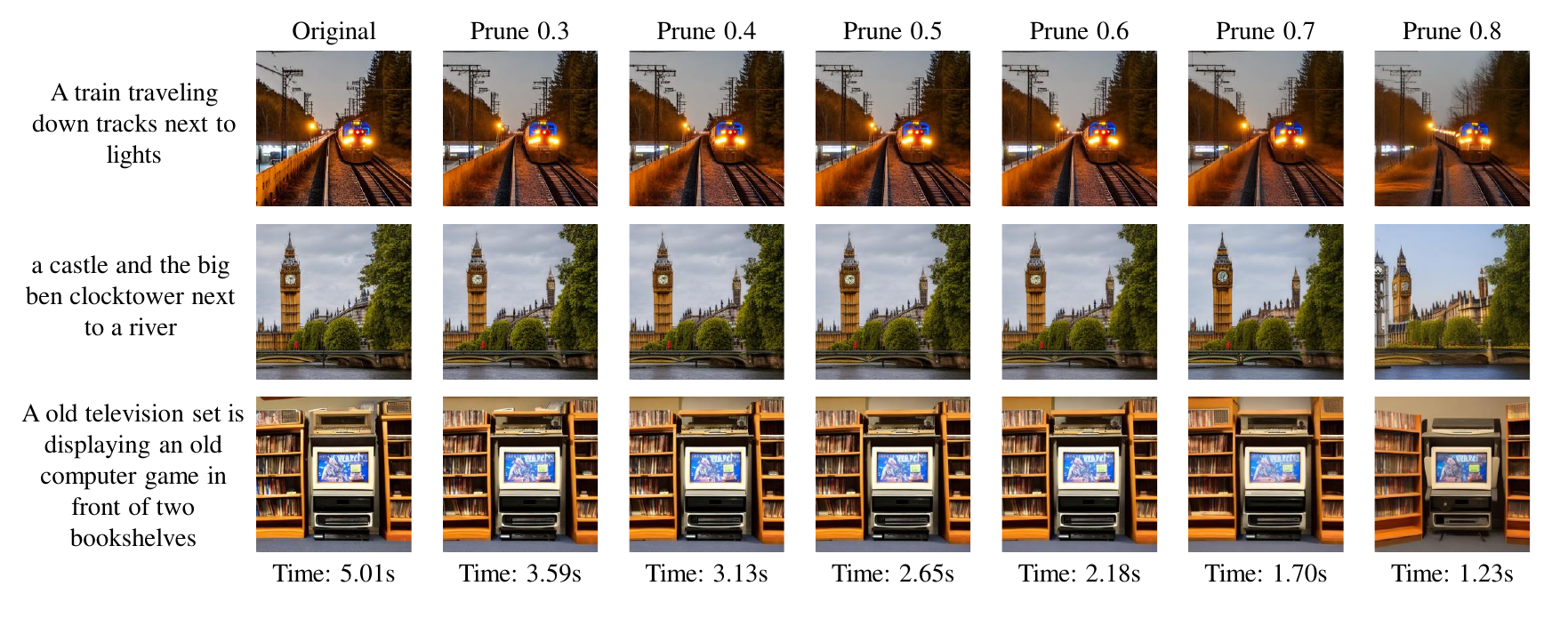}}
\caption{Visualization of generated images. It shows evolving patterns as pruning ratios increase. Despite these changes, main objects in the images remain consistent with the textual conditions.}
\label{fig:visualize}
\end{center}
\vspace{-6mm}
\end{figure*}

\paragraph{Qualitative Analysis of Increased Prune Ratio.}
In Figure \ref{fig:visualize}, we visualize the generated images as we increase the pruning ratio. With the increase in pruning ratio, the model's inference speed significantly improves, allowing us to achieve up to a fourfold increase in inference speed. However, as the pruning ratio increases, some patterns in the image content deviate from those in the original images. Nevertheless, our main objects in the figures consistently adhere to the textual conditions. Subtle changes in background details typically do not compromise image quality, as quantitatively analyzed in Table \ref{tbl:main_stable_diffusion}.

\section{Conclusion}
\vspace{-10pt}
This work explores resolving diffusion accelerating tasks by reducing redundant feature calculations across adjacent timesteps. We present a novel diffusion pruning framework and cast the model pruning process as a SubNet search problem. Our approach introduces a plugin pruner network that identifies an optimal SubNet through few-step gradient optimization. Results on a wide range of Stable Diffusion (SD) and DiT series models verify the effectiveness of our method. We achieve a 4.4$\times$ speedup on Stable Diffusion 1.5 and effectively prune the DiT model with few step optimizations.

\bibliographystyle{unsrt}
\bibliography{ref}

\clearpage
\appendix
\section{Memory Optimization Details}
\label{Append_A}
\textbf{Gradient Checkpointing.}
Due to the multi-step Markovian nature of sampling in diffusion models, updating the entire sampling process using gradient accumulation incurs significant memory costs, even with a batch size of 1. To mitigate this issue, we employ gradient checkpointing and half-precision floating-point training to reduce memory consumption. 
The core idea behind gradient checkpointing is to selectively preserve a portion of activation values during forward propagation, discarding the rest. During backpropagation, the gradients of the discarded activation values are computed using the saved gradients of the preserved nodes, effectively reducing memory usage. 
Additionally, we use gradient accumulation, wherein gradients computed over multiple iterations are accumulated and then backpropagated in a single batch for parameter updates, thus allowing for larger batch sizes under limited memory usage.

\section{Pseudo Code}
Here, we show the details of our proposed post-process algorithm via pseudo code as followings.
\label{Append_B}
\algdef{SE}[SUBALG]{Indent}{EndIndent}{}{\algorithmicend\ }%
\algtext*{Indent}
\algtext*{EndIndent}
\definecolor{myblue}{rgb}{0.7,0.7,0.7}
\algnewcommand{\LineComment}[1]{\State \textcolor{gray}{\(\triangleright\) #1}}

\begin{algorithm}[ht]
    \caption{Diffusion Pruner}
    \label{alg:post-p}
    \begin{flushleft}
        \textbf{Input:} A pretrained diffusion model $M$, importance scores $S$, a pruning ratio $p$  \\
        \textbf{Output:} The pruned diffusion model $M^*$
    \end{flushleft}
    \begin{algorithmic}[1]
        \State $\text{left} \gets 0.0$
        \State $\text{right} \gets 1.0$
        \While{$\text{True}$} 
            \State $\text{current} \gets (\text{left} + \text{right}) / 2$
            \State $S^* \gets S$
            \For{$t$ in $[0, 1, 2, ..., T]$}
                \For{each block score $s$ in $S^*$}
                    \If{$s < \text{current}$}
                        \State $s \gets 0$
                    \EndIf
                \EndFor
            \EndFor
            \State \text{update\_scores\_of\_blocks}  ($S^*$) // remove dependent blocks to set them zeros.
            \State $p^*, M^* \gets \text{prune\_diffusion\_model}(S^*, M)$ // obtain the pruned ratio and the pruned model.
            \If{$\text{abs}(p^* - p) < 0.0125$}
                \State \textbf{break}
            \ElsIf{$p^* < p$}
                \State $\text{left} \gets \text{current}$
            \Else
                \State $\text{right} \gets \text{current}$
            \EndIf
            \State $\mathcal{T} \gets \mathcal{T} / 2$
        \EndWhile
        \State \textbf{return} $M^*$
    \end{algorithmic}
\end{algorithm}

\newpage

\section{Additional Experiment Results}
\label{Append_C}
\begin{figure*}[ht]
\begin{center}
\centerline{\includegraphics[width=1.0\linewidth]{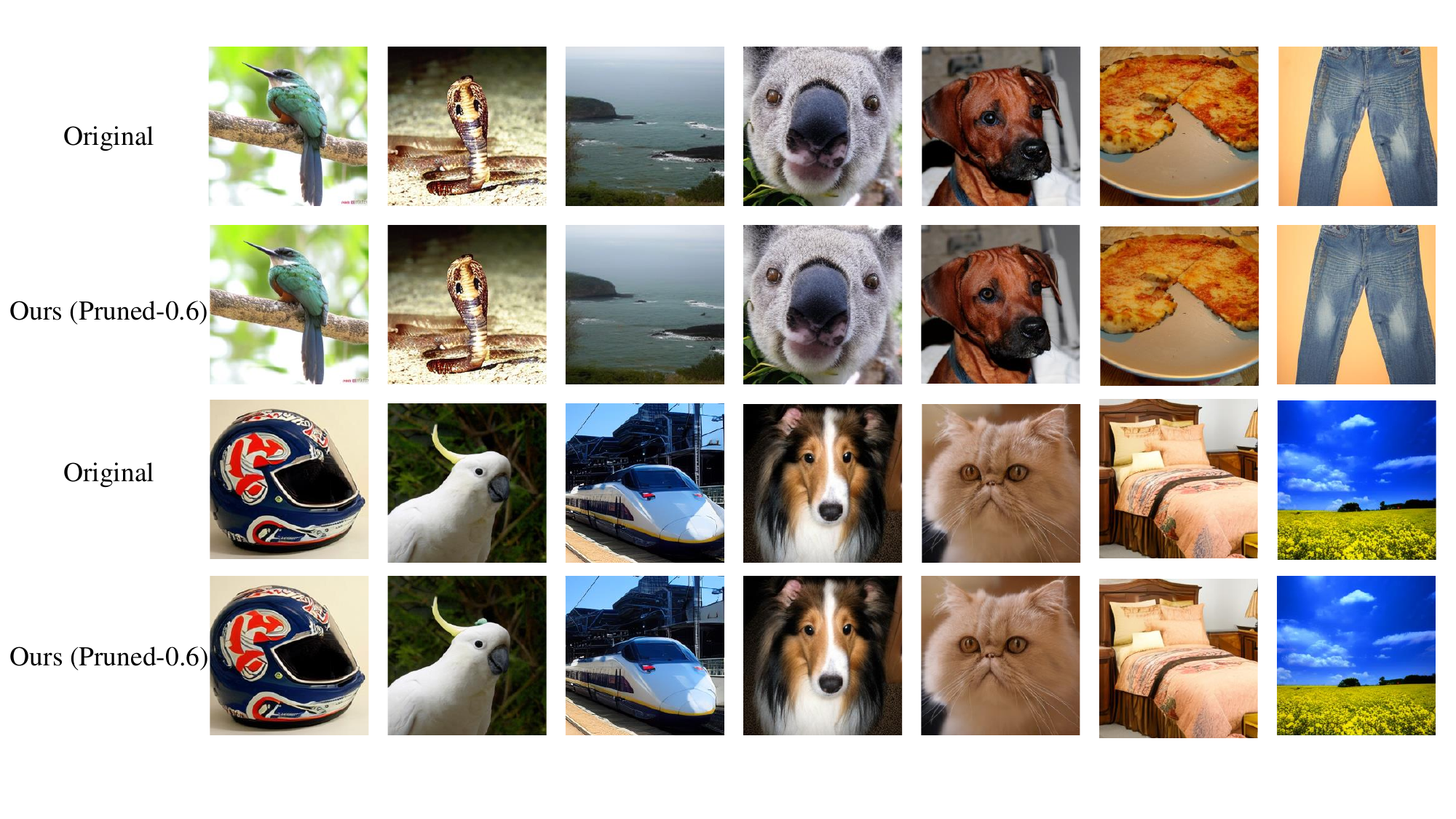}}
\caption{Visualization of DiT model generated images: samples using DDIM-250 steps (uplink) and pruned 60\% MACs (downlink). The speedup ratio here is $2.4 \times$ .}
\label{fig:results_compare_dit}
\end{center}
\vspace{-6mm}
\end{figure*}

\subsection{More Qualitative Results}
\textbf{Comparison with DiT Baselines.}
We provide the original unpruned DiT model and a version pruned by 0.6 ratio to generate comparison images  in Figure \ref{fig:results_compare_dit}. It can be observed that the plots generated by the pruned model are almost identical to those produced by the original model. Although there are slight differences in details, such as the appearance of the dog's eyes, these do not significantly affect the overall image quality.

\begin{figure*}[h]
\begin{center}
\centerline{\includegraphics[width=0.75\linewidth]{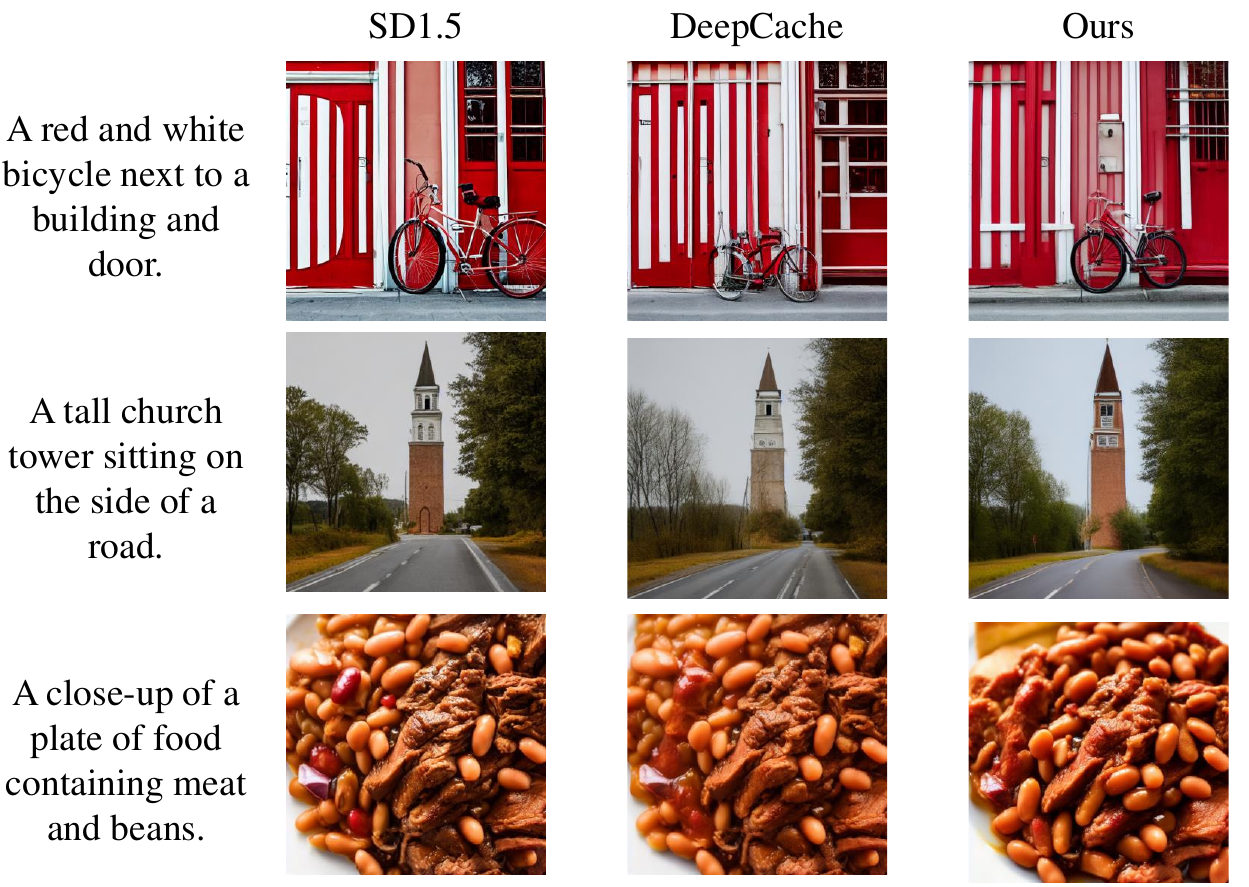}}
\caption{A qualitative comparison with existing methods is provided. We compare our method (prune 0.75) with DeepCache (N=4).}
\label{fig:vis_sd}
\end{center}
\end{figure*}

\textbf{Comparison with SD Baselines.} We provide qualitative comparisons with the SD baseline and DeepCache, as shown in Figure \ref{fig:vis_sd}. Our method demonstrates superior image-text consistency and image quality compared to existing methods.

\begin{figure*}[ht]
\begin{center}
\centerline{\includegraphics[width=0.75\linewidth]{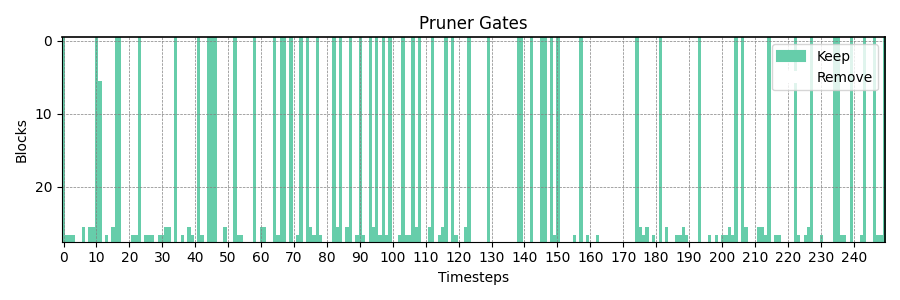}}
\caption{The visualization results of the pruning gates for DiT-XL/2 at 250 steps with a pruning ratio of 0.75.}
\label{fig:gate_vis}
\end{center}
\vspace{-8mm}
\end{figure*}

\textbf{Pruning Gate Visualization.}
Our method exhibits a specific pattern of pruning ratios with respect to the timesteps. As shown in Figure \ref{fig:gate_vis}, fewer blocks are pruned during the middle denoising stage (approximately between steps 65 and 150), as this is when the image content is rapidly being generated. Conversely, the pruning ratio in the latter stage is higher since the content has already taken shape.

\begin{figure*}[h]
    \centering
    \begin{minipage}{0.4\textwidth}
        \centering
        \includegraphics[width=\linewidth]{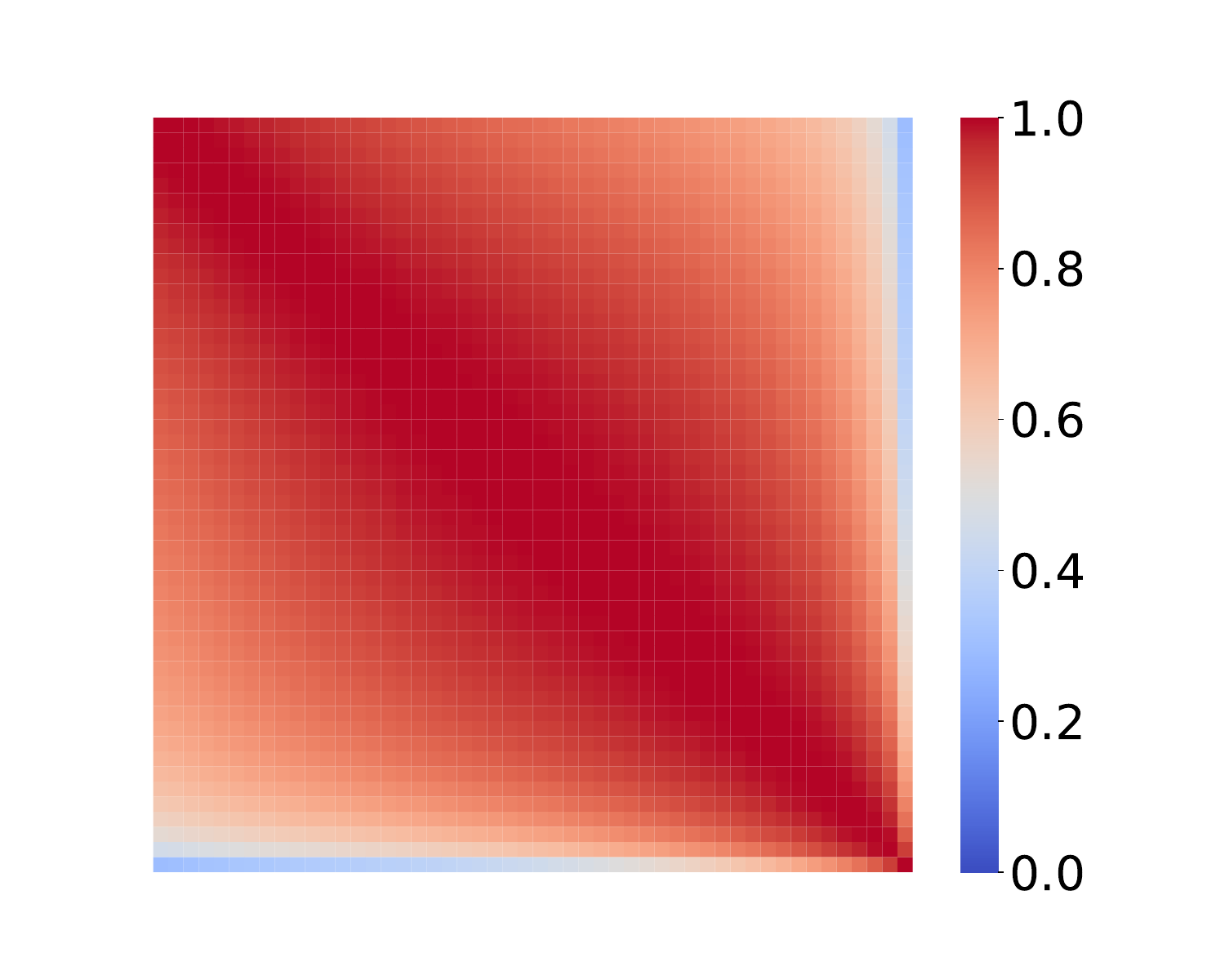}
        \small{(a) DDIM (SD-1.5 w/ 50 steps)}
        \label{fig:subfig1}
    \end{minipage}\hfill
    \begin{minipage}{0.4\textwidth}
        \centering
        \includegraphics[width=\linewidth]{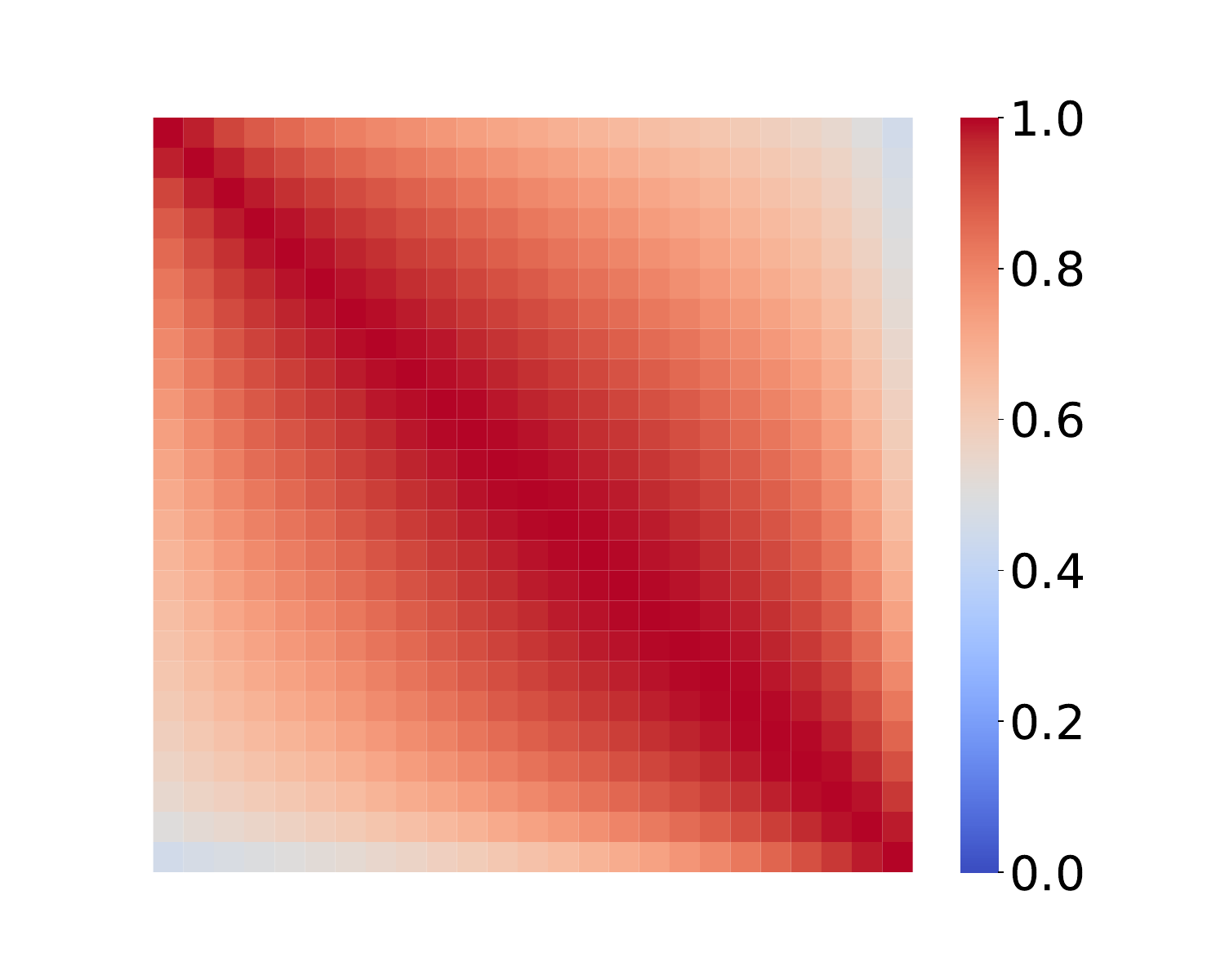}
        \small{(b) DPM (SD-2.1 w/ 25 steps)}
        \label{fig:subfig2}
    \end{minipage}
    \caption{Feature similarity across adjacent time steps in fast samplers.}
    \label{fig:sim_vis}
\end{figure*}

\textbf{Feature Similarity Analysis.} Recent studies have confirmed feature similarity across adjacent time steps \cite{ma2023deepcache, tgate}. We also conducted an analysis of feature similarity between adjacent steps in fast samplers. Specifically, we sampled 200 images from the COCO2017 validation set and calculated the average cosine similarity between the features of the penultimate upsampling block across all steps for two typical fast samplers, resulting in a similarity matrix, shown in Figure \ref{fig:sim_vis}. The heatmap in Figure \ref{fig:sim_vis} highlights the high degree of similarity between features at consecutive time steps.

\subsection{More Quantitative Results}

\textbf{More Ablations.}
We conducted an ablation study on $\alpha$ and present the results  in Table \ref{tab:alpha}. Our method achieves the best performance when $\alpha=1.0$.  we also conducted an ablation study on $\gamma$. Without $\gamma$, pruning 80\% on SD1.5 resulted in a CLIP score of 29.50 (w/ $\gamma$: 30.29).

\begin{table*}[ht]
\centering
\caption{Comparison of different $\alpha$ values. Pruning experiments with 80\% pruning ratio were conducted on COCO2017 validation using SD-1.5.}
\vspace{10pt}
\begin{tabular}{c|cccc}
\toprule
$\alpha$   & 0.1 & 0.5 & 1.0 & 2.0 \\\midrule
CLIP Score$\uparrow$  & 29.77  & 29.93  & \bf 30.29    & 30.17          \\ \bottomrule
\end{tabular}
\label{tab:alpha}
\end{table*}

\begin{table*}[!h]
\centering
\caption{Comparison with a 20-step DPM-Solver sampler for diffusion transformer model. We evaluate the effectiveness of our methods on COCO2017 validation set.}
\vspace{10pt}
\begin{tabular}{lccc}
\toprule
\textbf{Method $\downarrow$} & \textbf{MACs $\downarrow$} & \textbf{Speedup $\uparrow$} & \textbf{CLIP Score $\uparrow$} \\ \midrule
PixArt-$\alpha$ w/ 20-step DPM   & 85.65 T & 1.0 $\times$ & 30.43 \\
Pruned-0.4 (Ours)  & 51.39 T & 1.6 $\times$ & 30.41\\ \bottomrule
\end{tabular}
\label{tab:pixart}
\end{table*}

\textbf{More Baseline Comparison.} We also evaluated our method on PixArt-$\alpha$ \cite{chen2023pixart}, achieving excellent pruning results, as shown in Table  \ref{tab:pixart} below. Our method exhibits minimal performance loss on PixArt-$\alpha$ with a 0.4 pruning ratio.

\section{Limitations}
\label{Append_D}
A limitation of our method arises from its training process of the pruner network. Our method necessitates tuning an additional pruner network for the pre-trained diffusion model. This may entail users investing additional time when adapting our method to specific diffusion models. For example, we train DiP-GO for SD-1.5 on a single AMD Instinct MI250 GPU for $\sim$ 2.5 hours. However, we note that the introduced time is small compared to training a lightweight diffusion model. Besides, same as existing work, our method struggles to maintain performance with extremely high pruning ratios, presenting a challenge for deploying diffusion models in scenarios with severely limited computational resources. 

\section{Social Impact} 
Generative models have demonstrated promising results in content generation \cite{podell2023sdxl, chen2023pixart, rombach2022high}. However, due to the high inference costs, current methods struggle to achieve rapid application and deployment. Our approach introduces an efficient acceleration method for diffusion models, enabling nearly lossless speedup. Moreover, our method does not require retraining of the pretrained models and is compatible with various diffusion models, making it highly generalizable. This makes it suitable for rapid deployment of generative models on mobile and edge devices.

Nevertheless, since generative models are pretrained on large-scale internet datasets, the data they generate may contain inherent social biases and stereotypes \cite{cho2023dall, naik2023social, ross2020measuring}. Additionally, there is a risk of misuse, such as in the creation of DeepFakes \cite{lyu2020deepfake}, which could pose significant social harm. While reducing the usage cost, it is crucial to prevent the low-cost generative models from being misused, leading to negative societal impacts. Therefore, it is necessary to establish relevant laws and regulations, create a well-regulated community environment, and provide guidelines to ensure responsible dissemination and use of generative models.

\clearpage
\section*{NeurIPS Paper Checklist}

\begin{enumerate}

\item {\bf Claims}
    \item[] Question: Do the main claims made in the abstract and introduction accurately reflect the paper's contributions and scope?
    \item[] Answer: \answerYes{} 
    \item[] Justification: The abstract provides a concise summary of the main findings and contributions of the paper, while the introduction elaborates on the problem statement and research objectives, thereby clarifying the contributions.

\item {\bf Limitations}
    \item[] Question: Does the paper discuss the limitations of the work performed by the authors?
    \item[] Answer: \answerYes{} 
    \item[] Justification: In Limitation Section in Appendix~\ref{Append_D}, we expound upon the limitations of the work conducted and provide a brief discussion thereof.

\item {\bf Theory Assumptions and Proofs}
    \item[] Question: For each theoretical result, does the paper provide the full set of assumptions and a complete (and correct) proof?
    \item[] Answer: 
    \answerNo{}
    \item[] Justification: None.

    \item {\bf Experimental Result Reproducibility}
    \item[] Question: Does the paper fully disclose all the information needed to reproduce the main experimental results of the paper to the extent that it affects the main claims and/or conclusions of the paper (regardless of whether the code and data are provided or not)?
    \item[] Answer: \answerYes{} 
    \item[] Justification: In Section \ref{sec: Experimental setup}, we introduced the details of experimental setup and model training to ensure reproducibility.

\item {\bf Open access to data and code}
    \item[] Question: Does the paper provide open access to the data and code, with sufficient instructions to faithfully reproduce the main experimental results, as described in supplemental material?
    \item[] Answer: \answerNo{} 
    \item[] Justification: The code will be released after it successfully passes our company's internal review.

\item {\bf Experimental Setting/Details}
    \item[] Question: Does the paper specify all the training and test details (e.g., data splits, hyperparameters, how they were chosen, type of optimizer, etc.) necessary to understand the results?
    \item[] Answer: \answerYes{} 
    \item[] Justification: In Section \ref{sec: Experimental setup}, we introduced the details of experimental setup and model training and testing. 

\item {\bf Experiment Statistical Significance}
    \item[] Question: Does the paper report error bars suitably and correctly defined or other appropriate information about the statistical significance of the experiments?
    \item[] Answer: \answerNo{} 
    \item[] Justification: The experiments conducted in our paper do not involve the use of error bars or statistical significance analysis, thus this aspect is not applicable to our study. 

\item {\bf Experiments Compute Resources}
    \item[] Question: For each experiment, does the paper provide sufficient information on the computer resources (type of compute workers, memory, time of execution) needed to reproduce the experiments?
    \item[] Answer: \answerYes{} 
    \item[] Justification: For our experiments, we furnished detailed specifications of the GPU models used along with their corresponding tasks. Furthermore, we included specific information regarding the model training batch size and the number of training iterations.
    
\item {\bf Code Of Ethics}
    \item[] Question: Does the research conducted in the paper conform, in every respect, with the NeurIPS Code of Ethics \url{https://neurips.cc/public/EthicsGuidelines}?
    \item[] Answer: \answerYes{} 
    \item[] Justification: We have carefully reviewed the NeurIPS Code of Ethics and adhere to its principles.

\item {\bf Broader Impacts}
    \item[] Question: Does the paper discuss both potential positive societal impacts and negative societal impacts of the work performed?
    \item[] Answer: \answerYes{} 
    \item[] Justification: We discuss both potential positive societal impacts and negative societal impacts of the work performed in Appendix.
    
\item {\bf Safeguards}
    \item[] Question: Does the paper describe safeguards that have been put in place for responsible release of data or models that have a high risk for misuse (e.g., pretrained language models, image generators, or scraped datasets)?
    \item[] Answer: \answerNA{} 
    \item[] Justification: Our paper poses no such risks.

\item {\bf Licenses for existing assets}
    \item[] Question: Are the creators or original owners of assets (e.g., code, data, models), used in the paper, properly credited and are the license and terms of use explicitly mentioned and properly respected?
    \item[] Answer: \answerYes{} 
    \item[] Justification: The creators or original owners of assets, such as code, data, or models, used in the paper, are properly credited. Additionally, the license and terms of use associated with these assets are explicitly mentioned and respected in accordance with ethical and legal standards.

\item {\bf New Assets}
    \item[] Question: Are new assets introduced in the paper well documented and is the documentation provided alongside the assets?
    \item[] Answer: \answerNA{} 
    \item[] Justification: Our paper does not release new assets.

\item {\bf Crowdsourcing and Research with Human Subjects}
    \item[] Question: For crowdsourcing experiments and research with human subjects, does the paper include the full text of instructions given to participants and screenshots, if applicable, as well as details about compensation (if any)? 
    \item[] Answer: \answerNA{} 
    \item[] Justification: Our paper does not involve crowdsourcing nor research with human subjects.

\item {\bf Institutional Review Board (IRB) Approvals or Equivalent for Research with Human Subjects}
    \item[] Question: Does the paper describe potential risks incurred by study participants, whether such risks were disclosed to the subjects, and whether Institutional Review Board (IRB) approvals (or an equivalent approval/review based on the requirements of your country or institution) were obtained?
    \item[] Answer: \answerNA{} 
    \item[] Justification: Our paper does not involve crowdsourcing nor research with human subjects.

\end{enumerate}

\end{document}